\UseRawInputEncoding
\documentclass[letterpaper, 10 pt, conference]{ieeeconf}  

\IEEEoverridecommandlockouts                              

\usepackage{graphics} 
\usepackage{epsfig} 
\usepackage{mathptmx} 
\usepackage{times} 
\usepackage{amsmath} 
\usepackage{amssymb}  

\usepackage{caption}
\usepackage{subcaption}
\usepackage{booktabs}
\usepackage{hyperref}

\usepackage{microtype}
\usepackage{colortbl}
\usepackage{xcolor}
\usepackage{float}
\usepackage{float}
\usepackage{cuted}
\usepackage{capt-of}
\usepackage{url}

\title{\LARGE \bf
Real-time Puncture Detection and Recovery for Pneumatic Soft Actuators
}

\author{Tejonidhi R. Deshpande $^{1}$, Tingyu Cheng$^{2}$ and Josiah Hester$^{1}$
\thanks{$^{1}$Tejonidhi and Josiah are with the School of Interactive Computing, College of Computing
        Georgia Institute of Technology, Atlanta, GA,USA
        {\tt\small tdeshpande33@gatech.edu, josiah@gatech.edu}}%
\thanks{$^{2}$Tingyu is with the Department of Computer Science and Engineering,
        University of Notre Dame, St. Joseph County, IN, USA
        {\tt\small tcheng2@nd.edu}}%
\thanks{Research was sponsored by the Army Research Office and was accomplished under Cooperative Agreement Number W911NF-23-2-0138. The views and conclusions contained in this document are those of the authors and should not be interpreted as representing the official policies, either expressed or implied, of the Army Research Office or the U.S. Government. The U.S. Government is authorized to reproduce and distribute reprints for Government purposes notwithstanding any copyright notation herein.}
}

\begin{document}

\maketitle
\thispagestyle{empty}
\pagestyle{empty}

\begin{abstract}

Soft robots offer safe and adaptive interaction with humans and unstructured environments through their inherent ability to deform and comply. Pneumatic actuators are one way to build soft robots. They are typically made from soft silicone materials and are especially effective for driving such systems, enabling smooth and adaptable motion. However, their compliant nature also makes them vulnerable to mechanical failures like punctures and tears, limiting practical deployment. To address this, we propose a puncture detection system for soft actuators using motion data from a single inertial measurement unit. Extracted features are used to train anomaly detectors for puncture detection and non-linear models to estimate severity. We also introduce a multi-chamber pneumatic soft bending actuator capable of diverse configurations via selective chamber inflation. Our algorithm identifies the punctured chamber and provides a severity score using a chamber perturbation scheme. Anomaly detectors are trained on normal operation data and detect damage through reconstruction errors, while severity is estimated by a separate model trained under slightly modified conditions. Finally, we demonstrate a failure recovery strategy to maintain actuation force post-failure. This approach enhances the reliability and safety of soft robotic systems through real-time, data-driven damage detection.

\end{abstract}

\section{INTRODUCTION}
Soft robots, built from compliant materials and structures, are uniquely suited for safe interaction with fragile bodies and exhibit robust performance in unstructured environments. Their flexibility enables them to absorb impacts, conform to complex surfaces, and navigate constrained spaces that traditional rigid robots would struggle to operate in. Among them, pneumatic soft robots occupy a central role: lightweight, efficient, and capable of large deformations driven by pressurized air, they can be fabricated via mold casting or 3D printing using low-cost elastomers such as silicone \cite{SHAPIRO2011484}. Pneumatic actuators—whether chambered networks \cite{8693775} or fiber-reinforced artificial muscles\cite{PneumaticMuscle} —offer broad motion capabilities and inherent safety, supporting applications in underwater exploration \cite{li2021self}, medical rehabilitation \cite{miyashita2016ingestible}, wearable assistance \cite{ yin2021wearable}, and industrial automation \cite{9144922}.

 \begin{figure}[thpb]
      \centering
      \includegraphics[width=1\linewidth]{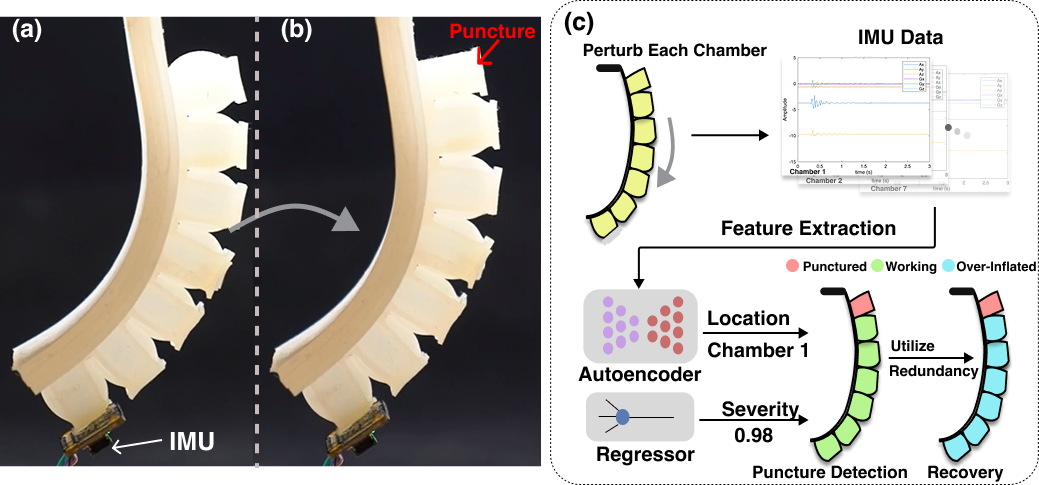}
      \caption{Illustration of puncture detection scheme for soft-robots using chamber perturbation. (a) Demonstrates the multi-chamber soft actuator configuration with all chambers working (b) Demonstrates the actuator configuration when 1st chamber fails. (c) Illustrates the proposed puncture detection and recovery pipeline.}
      \label{fig:overview}
   \end{figure}

 Although the flexibility and compliance offer advantages for the soft robots made from ultra-soft materials such as Smooth-On 00-50, they also present challenges, particularly concerning the structural integrity of the system. Anomalies such as air leakage or material puncture can lead to critical failures thereby hampering performance and reduce safety~\cite{Abidi2017}. 
 These challenges are further exacerbated when the soft actuators are made from less resilient materials, such as biodegradable/bioabsorbable materials (e.g, gelatin, hydrogel) which due to the material's own property, are inherently more susceptible to damage. 
 Therefore, to enhance the resilience of soft actuators, it is essential to implement effective puncture detection and mitigation (or recovery) strategies that monitor the robot's integrity and functionality in real-world applications, thereby enabling reliable performance under uncertain conditions, reducing downtime, and extending the operational lifespan of soft-robotic systems.

 In this research, we develop a robust real-time puncture-detection algorithm for pneumatically driven soft actuators that can identify the location and severity of a punctured chamber in our multi-chamber pneumatic-bending actuator design. A puncture in the soft actuator's chamber causes changes in the inflation and deflation motion patterns of the soft actuators. The proposed localization algorithm captures these differences and flags any deviations above a threshold as an anomaly or puncture, while the severity model outputs a score between 0 (low severity) and 1 (high severity). The algorithm achieves over 96\% accuracy in localizing the punctured chamber using time series data from a single inertial measurement unit (IMU). 
 This allows for a nuanced understanding of the system’s operational state, improving both diagnostic precision and system reliability by enabling targeted maintenance by identifying the puncture location and corresponding response strategies. 
 Our multi-chamber pneumatic bending actuator design (see Fig.~\ref{fig:overview}) offers mechanical redundancy—allowing continued operation even in the presence of a chamber failure by leveraging deformation from intact chambers.

This paper presents the first puncture-detection algorithm capable of operating with only one compact and low-power onboard sensor —an IMU with 6 degrees-of-freedom —making it highly suitable for untethered soft robots in field applications, which often have payload constraints. The key contributions of this work are as follows:
\begin{enumerate}
    \item Development of a lightweight puncture detection algorithm that identifies the location of the punctured chamber, along with severity, and relies solely on onboard IMU data.
    \item A novel perturbation scheme enabling puncture localization and severity scoring.
    \item A novel recovery scheme allowing the soft-actuator to retain its tip-force even after puncture.
    \item Release of a comprehensive database of IMU readings corresponding to punctures in different chambers, specifically for a bending silicone-based actuator\footnote{Dataset: \href{https://github.com/ka-moamoa/softbot-punctures}{https://github.com/ka-moamoa/softbot-punctures}}
\end{enumerate}

\section{Related Work}

\subsection{Puncture Detection for Soft Actuators}

Punctures in soft actuators often lead to significant, undesirable model changes and can be considered anomalies \cite{Abidi2017}. Anomaly or fault detection refers to identifying failures in a monitored system or process based on its current state when it deviates from its normal operation \cite{s22062205}. While anomaly detection has extensively studied for industrial processes and rigid robots, anomaly detection for soft robots remain largely unexplored. The deformability and inherent compliance of these systems pose unique challenges for sensing, modeling, and experimental repeatability. Based on the underlying learning paradigm, anomaly detection methods can be classified into supervised, semi-supervised, and unsupervised approaches. 

The supervised anomaly detection method requires data from both normal and anomalous classes \cite{Jimultimodal}. Still, its wide applicability is limited in robotics as the availability of anomalous data is usually limited. Unsupervised anomaly detection, in contrast, does not require any labeled training data. Such methods often employ clustering methods, such as k-means \cite{9072123} and distance-based clustering to learn patterns in the data \cite{8101921}. Data points that deviate significantly from these patterns are flagged as outliers. While the lack of labeled data makes these approaches more flexible, it also introduces a major challenge: without ground truth, it becomes difficult to assess the model’s performance. This often leads to a high false-positive rate, particularly in complex scenarios such as robotic systems interacting with dynamic environments. \cite{alvarez2022revealinglargescaleevaluationunsupervised}

Semi-supervised methods offer a middle ground, in which a one-class classifier is often trained only on normal (non-anomalous) data. Any data that deviates from the learned distribution is flagged as anomalous. Some candidate examples for such classifiers are one-class support vector machines (OC-SVM) 
\cite{Barbado_2022}, k-means based detectors, isolation forests\cite{9617498}, etc. Recent advances in semi-supervised anomaly detection field have utilized a reconstruction-based approach that uses autoencoders to compress and reconstruct high-dimensional information from non-anomalous data inputs. As the autoencoder is trained only on non- anomalous data, an anomaly can be indicated with a high reconstruction error, as the autoencoder struggles to accurately reconstruct unfamiliar anomalous patterns \cite{8279425}.

In this paper, we utilize a fully connected network based autoencoder for multi-modal puncture localization in pneumatic soft robots for their simplicity, interpretability and efficiency on smaller datasets. To calculate the severity score, we use a single perceptron with tanh activation.

\begin{figure}[thpb]
      \centering
      \includegraphics[width=1\linewidth]{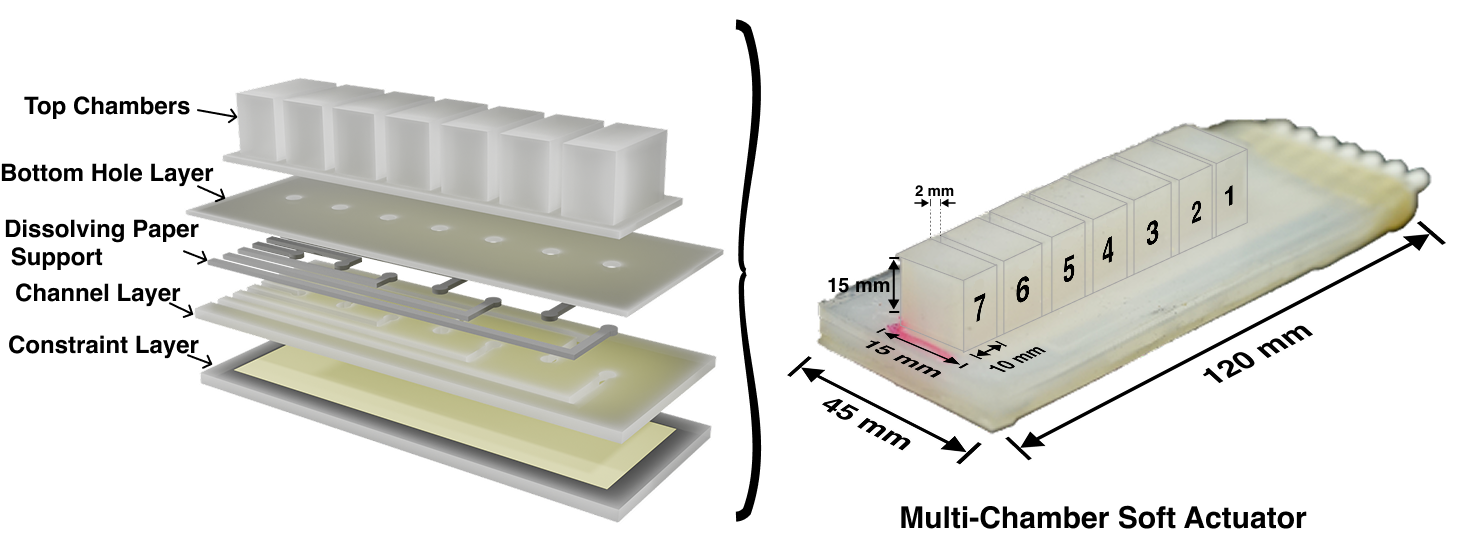}
      \caption{Multi-chamber soft-actuator design with chamber numbers.}
      \label{fig:fabrication}
   \end{figure}

\subsection{Sensing for failure detection}

Sensing in soft robotics has advanced significantly in recent years, with researchers exploring a diverse range of modalities to monitor deformation, internal pressure, and external forces. Common approaches include embedded strain sensors \cite{s140712748}
, capacitive \cite{mannsfeld2010highly}, resistive pressure sensors \cite{pan2014ultra}, optical fiber-based sensors \cite{ramuz2012transparent}, and liquid metal-based sensors \cite{11128570}. These sensors are typically integrated directly into the soft robot body or actuator to enable real-time measurement of physical quantities during operation. For instance, Krings et al. proposed an autonomous damage-detection approach by embedding liquid-metal droplets within a silicone elastomer matrix \cite{11128570}. While embedded sensing enables high-fidelity monitoring, it often introduces additional design complexity, and may compromise the mechanical compliance and softness that are central to the performance of soft robotic systems.

Some of the other approaches use indirect estimation \cite{thuruthel_soft_2019}. developed a method for modeling soft actuated systems using vision-based motion capture system and recurrent neural nets. In addition to custom, lab-fabricated sensors, commercially available sensors such as IMUs have also been employed for direct sensing in soft robots. IMUs offer a low-cost, compact, and efficient solution for enabling precise control, real-time state estimation, and safe operation of soft pneumatic actuators—without compromising the compliance and softness of the system \cite{7139569, 8624340}.

A more straightforward method for detecting punctures would be to perform pressure-based sensing for each actuator chamber, where significant and frequent undesired pressure drops can indicate a puncture. However, this approach significantly increases both system cost and integration complexity. Moreover, pressure readings are often affected by the non-linear and hysteretic behavior of soft materials, which can limit their ability to accurately capture dynamic changes\cite {Qu2022}.

In contrast, IMUs can indirectly indicate puncture-related failures by detecting deviations in the expected motion response of the soft robot. These deviations reflect changes in internal pressure or actuation efficiency caused by a puncture. Due to their versatility and effectiveness, IMUs are chosen in this work as the primary and only sensing modality for puncture detection.

\section{System Overview}

\subsection{Multi-Chamber Soft Actuator Design}

In the scope of this work, we designed a pneumatic bending multi-chamber soft-actuator emphasizing on redundancy and reliability, in which each pneumatic chamber can be inflated or deflated individually. The motivation behind such a multi-chamber design is to introduce a degree of redundancy in the system such that if one or more chamber is punctured, the remaining functional chambers can continue operation preventing a total failure. This design features internal air channels that lead to each chamber, as shown in Fig. \ref{fig:fabrication}. 

\subsection{Experimental Test Bed Setup} 

\begin{figure}[t]
      \centering
      \includegraphics[width=1\linewidth]{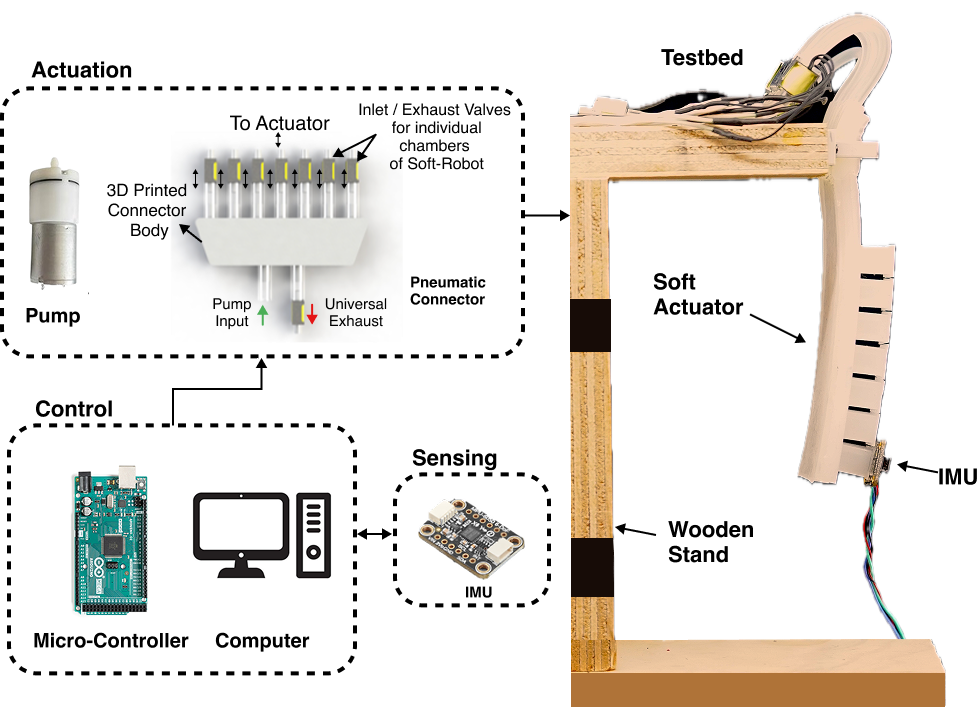}
      \caption{Overview of the testbed used for data collection.  }
      \label{fig:testbed}
   \end{figure}

To enable efficient data collection, we developed a custom test bed as shown in Fig. \ref{fig:testbed}. The test bed consists of a wooden stand for mounting the soft actuator, pneumatic connections, control electronics, and a portable pneumatic pump. The suspended design is for data collection, ensuring reduction in uncertainties such as friction and surface interactions, making the actuator motion reliable and repeatable.
 
The pneumatic connector features multiple pneumatic solenoid valves (DC4.5V 0.5A, 2 Way Normally Closed) to control air inlet and exhaust to individual chambers. It serves as the interface between the soft-robot and the pump. The control electronics includes Arduino Mega 2560 which acts as a low level interface, receiving actuation commands from PC. A compact 9V pneumatic pump (370 air pump, 3 L/min) is used as a source of pressurized air. A 3mm soft-silicone tubing is used to facilitate connections between the pump, the connector and the actuator. The IMU (MPU-6050) mounted on the actuator is used to collect the motion data. 

\subsection{Data Collection for Puncture and Severity Detection}

\subsubsection{Data Collection for Puncture Localization}

To the best of our knowledge, no publicly available datasets exist for puncture detection in soft-actuators. Consequently, we collected a custom dataset specifically tailored for training and evaluating classifier models. Our puncture localization data comprises three datasets, described below. The data collection process for puncture localization involved pre-inflating all the chambers of the actuator to a nominal pressure to mimic the actuated state and perturbing each chamber for normal (non-anomalous) as well as puncture/fail  (anomalous) cases while collecting 6-axis IMU data. While the single-chamber dataset was used to train all models, the other two datasets (multi-chamber and tip-contact) were used to validate robustness of the model. 

\begin{itemize}
    \item Dataset 1 : Consists of IMU data corresponding to a single chamber perturbing while it is normally working, and when it is punctured. Approximately 100 data points were collected for each chamber working and puncture case constituting 1388 time series samples in total (682 working, 706 fail).

    \item Dataset 2: Consists of IMU data corresponding to chamber 4 perturbing when chamber 3 and chamber 5 have failed, mimicking a multi-chamber failure. 99 data points were collected (49 working, 50 fail) when chamber 4 is normally working and punctured. 

    \item Dataset 3: Data collection in this case followed similar process to that of Dataset 1 but in this case the actuator tip made consistent contact with a stationary object fixed to ground, significantly damping the oscillations in the actuator due to the perturbation. We collected 103 data points (53 working, 50 fail) for this case, using only chamber 3.

\end{itemize}

While Dataset 1 was used for model training, Datasets 2 and 3 were specifically collected to evaluate the model’s robustness under realistic operating conditions. In particular, these datasets incorporate commonly encountered but previously unseen variations—such as multi-chamber failures and contact-induced damping—that were not represented in the training data. This design enables a systematic assessment of the model’s ability to generalize beyond the conditions observed during training. Although not exhaustive, the selected scenarios capture the most frequent external disturbances and failure modes in soft robotic systems, thereby providing a practically meaningful and representative robustness evaluation.

The perturbation scheme used to collect data for a particular chamber consisted of a 20 ms pulse inflation, a 500 ms delay and 20 ms pulse deflation for the chamber under test. These values were selected empirically based on response times of the solenoid valves.

\subsubsection{Data Collection for Severity Score}

To quantify the severity, it is essential to determine how rapidly the pressure drops in a chamber. To gain an understanding of this, we collect data for different puncture sizes using the same perturbation scheme mentioned above, but instead of a single deflation pulse, 4 deflation pulses were used at the end. We collect 50 data points each for circular puncture sizes of  0.51 mm, 1.06 mm, 1.30 mm, and 2.00 mm, as well as no puncture case using this perturbation scheme. 

\begin{figure}[t]
      \centering

      \includegraphics[width=1\linewidth]{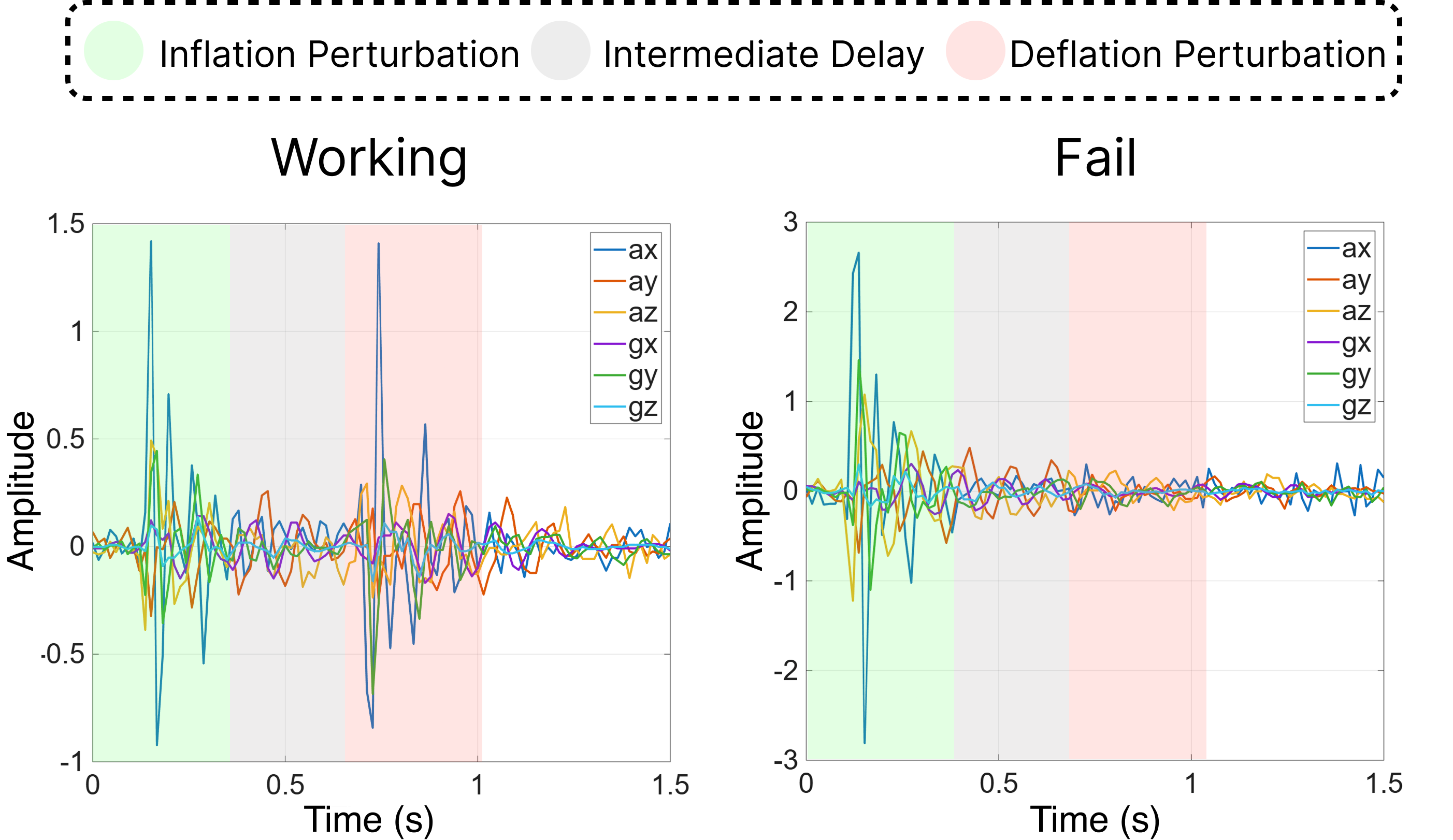}
      \caption{IMU data visualization for chamber 3 with the puncture localization perturbation scheme for both normal (working) and puncture (fail) case, with all other chambers functional and inflated.}
      \label{fig:DataViz}
   \end{figure}

The inflation-deflation perturbation scheme enables a nuanced assessment of individual chamber states while maintaining the overall actuator pose close to its initial configuration prior to perturbation. Introducing a deliberate delay between inflation and deflation phases helps to temporally separate their respective system responses, reducing noise caused by natural vibrations during pulse-inflation and allowing additional time for air leakage to manifest in the presence of a puncture. This separation enhances the robustness of puncture detection. Notably, the deflation response exhibited a marked difference between working and punctured chambers (see Fig. \ref{fig:DataViz}), which proved critical for accurate failure localization. This observation aligns with the physical intuition that further deflation of an already compromised (punctured) chamber produces minimal vibrations, thereby providing a distinctive signal for identifying failure.

\begin{figure*}[t]
      \centering
      \includegraphics[width=1\linewidth]{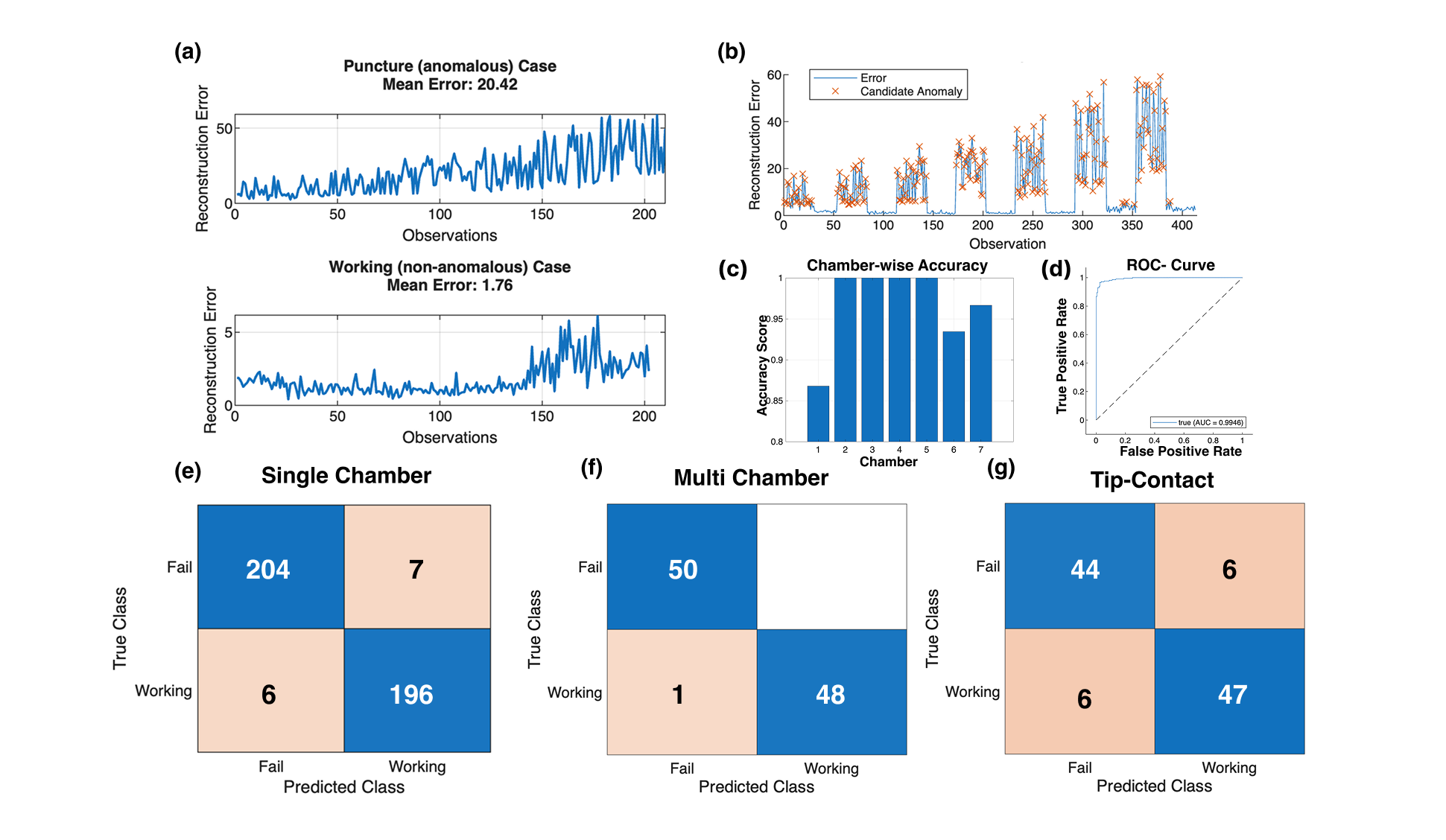}
      \caption{Punctured chamber localization results summary using FCN autoencoder. (a) Shows a comparison between test data reconstruction errors for working and fail input to the autoencoder for Dataset 1.  (b) Reconstruction error vs test data observations plot (c) Chamber-wise puncture detection accuracy for the single chamber case (d) Receiver operating characteristic curve for the FCN autoencoder-based puncture detection model. (e,f,g) confusion matrices for puncture detection for Datasets 1, 2, and 3, respectively.}
      \label{fig:Results Summary}
   \end{figure*}

\subsection{Feature Engineering}

 After filtering the segmented IMU data, we extract relevant signal-based features and use them to train the classifiers. The selected feature set comprises statistical features (mean, standard deviation, RMS, shape factor, kurtosis, and skewness) and impulsive features (crest factor, impulse factor, clearance factor, and peak value). These features summarize complex IMU data to capture meaningful patterns, particularly for time-series vibration data \cite{machines10060459, Fidali2024}. Extracting these features from each axis of the 6-axis IMU data (Ax, Ay, Az, Gx, Gy, Gz) resulted in 60 features in total per time series. These features are further ranked using t-test scores and top 20 features are selected to train and test the classifier for failure localization, as well as to fit a non-linear regression model for the severity score.

\subsection{Puncture Localization and Severity Score Algorithm}

Detecting punctures in pneumatic soft-robots is a multi-step process described in Fig.\ref{fig:overview}(c). To begin the puncture detection, all chambers are perturbed one by one while IMU data are collected. The IMU data are then segmented into 7 segments such that each segment contains IMU data corresponding to exactly one chamber.

To this end, we explored a variety of classifiers commonly employed for anomaly detection using time-series data. Specifically, we trained and tested one-class SVMs, isolation forests, and fully connected network (FCN) based auto-encoders of varying sizes as they are effective at capturing the statistical and temporal characteristics of normal operation, thereby enabling reliable detection of puncture-related deviations in soft actuators. Puncture detection with these models involved training the model only on working (non-anomalous) data. 

One-class SVM learns a compact decision boundary that encloses the majority of the working (non-anomalous) instances in the feature space, flagging any deviations as potential anomalies. In contrast, the isolation forest identifies anomalies by measuring how easily data points can be separated from the rest of the dataset. Since, anomalous points tend to be isolated in fewer random partitions, they receive higher anomaly scores. The FCN autoencoder model learns to reconstruct the input during training. Anomaly detection becomes possible using this reconstructed input as the autoencoder struggles to reconstruct the output correctly when the input corresponds to fail (anomalous) or puncture data, resulting in a large reconstruction error. In this way, the input data can then be classified as working or failure based on a pre-defined threshold on the reconstruction error. This threshold-based approach is useful for detecting various failures resulting in abnormal motion patterns in the actuator.

Severity scoring is formulated as a regression problem, where the objective is to predict a continuous value between 0 (indicating low severity) and 1 (indicating high severity) from observed input features. Analysis of the dataset revealed a non-linear relationship between pressure drop and puncture size (Fig. \ref{fig:Severity Compare}b), necessitating a non-linear modeling approach. To this end, we employ a single-layer perceptron with a hyperbolic tangent ($\tanh$) activation function, which effectively captures the non-linearity while maintaining model simplicity.

\section{Experimental Results}

\subsection{Localization of punctured chamber}

Puncture localization in pneumatic soft actuators is enabled by the distinct motion response observed during inflation and deflation when compared to intact actuators, as illustrated in Fig. \ref{fig:DataViz}. This section presents a comparative evaluation of three anomaly detection models, each designed to effectively capture the dynamic differences between punctured and non-punctured actuators for accurate fault detection in soft robotic systems. All the models were trained on working (non-anomalous) data from Dataset 1.

The most effective and reliable puncture detection approach involved using an autoencoder with fully connected networks, achieving 96.85 \% accuracy (see Fig. \ref{fig:Results Summary} e). The auto-encoder model was able to accurately reconstruct the input features for samples corresponding to the working case while generating large reconstruction errors for the fail case (\ref{fig:Results Summary} b).  e.g., during testing, the mean reconstruction error in feature space for fail data was 20.42, while that for working data was 1.76 (Fig. \ref{fig:Results Summary} a). All the samples corresponding to a reconstruction error of 40 \% or above ($thresh = 0.4$), that of the mean reconstruction error, were flagged as fail (anomaly), indicating puncture in the corresponding chamber. With 0.97 precision and 0.96 recall, this method proves to be promising while still minimizing false positives. 

We further evaluated the robustness of the trained classifier using Datasets 2 and 3. For Dataset 2, the model maintained a high accuracy of 99\% without any fine-tuning ($thresh = 0.4$) to the original model (Fig. \ref{fig:Results Summary} f) trained on Dataset 1. This high accuracy score was a result of using data collected from just 1 chamber and the model achieves higher accuracy if individual chambers are considered. (Fig.\ref{fig:Results Summary} c). 

In Dataset 3, the actuator tip made contact with a fixed object during motion. This contact-induced damping significantly altered the vibration profile, leading to a drop in classification accuracy to below 80\% using the same model and threshold. However, after fine-tuning the threshold ($thresh = 1.6$) it achieved 91\% accuracy (Fig. \ref{fig:Results Summary}, g). These results underscore the model's sensitivity to external damping effects and highlight the importance of contextual adaptation of the threshold for reliable performance.

We also evaluated the performance of unsupervised anomaly detection methods, specifically one-class SVMs and isolation forests. On Dataset 1, the one-class SVM achieved a high accuracy of 99.03\% with a respective precision and recall score of 1.0 and 0.98 , demonstrating strong discriminative capability. Similarly, the isolation forest model, implemented with an ensemble of 100 base learners, achieved an accuracy of 89.59\% (precision = 0.83, recall = 1.0).
While both one-class SVM and isolation forest models performed exceptionally well on data from Dataset 1, their generalization ability was limited. When tested on Dataset 2 and Dataset 3, the accuracy of both models dropped below 80\%. This performance degradation is likely due to changes in the input signal characteristics, particularly the damping effects introduced by adjacent chamber failures or environmental interactions, which were not present in the training data.

\begin{figure*}[t]
      \centering
      \includegraphics[width=1\linewidth]{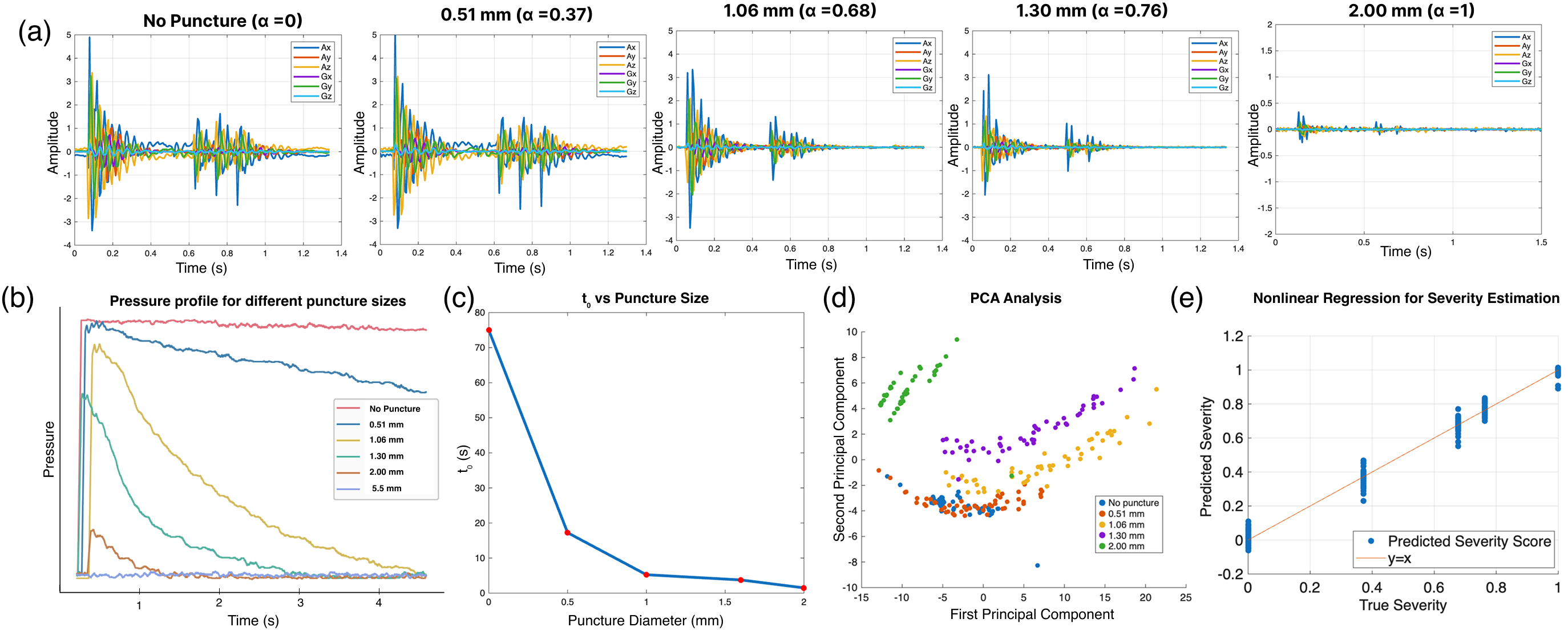}
      \caption{(a) IMU response data for severity chamber perturbation for different puncture diameter sizes and severity score $\alpha$ for chamber 3 perturbation, (b) Pressure profile for different puncture sizes, (c)Variation of ${t_0}$ with respect to puncture size (d) Scatter plot of first 2 principal components from extracted features (e) Non-linear regression analysis for puncture severity estimation.}
      \label{fig:Severity Compare}
\end{figure*}

\subsection{Puncture Severity Score}

Quantifying puncture severity is critical for enabling informed control decisions in the presence of damage. Chambers with minor punctures may retain sufficient functionality to continue operation without significantly impacting overall system performance. In contrast, chambers with severe punctures may be better excluded from actuation, allowing the system to leverage built-in redundancies to recover or maintain desired behavior. In this section we describe our approach to quantify severity using pressure data and develop a non-linear regression model fit on extracted features from IMU data to output a severity score between 0 and 1, where a higher score indicates more severe puncture and rapid pressure loss. 

We quantify puncture severity $\alpha$ for different puncture sizes as follows,

$$ \alpha = \frac{\log{(t_{max})} -\log{(t_{0})}}{\log{(t_{max})}-\log{(t_{min})}}$$

where, $t_0$ is the time required for the system to return to its initial pressure level after undergoing the same inflation cycle, and  $t_{max}$ and $t_{min}$ are the maximum and minimum values of $t_0$ from different puncture sizes we tried. This was particularly useful as a smaller $t_0$  is not linearly more severe. Fig. \ref{fig:Severity Compare} (b) demonstrates the pressure response of chamber 3 for different puncture sizes for the same inflation cycle, with (c) denoting $t_0$ values. We then use $\alpha$ to label the severity dataset. We consider no-puncture case having a severity score of 0 and a puncture with 2.00 mm size having a severity score of 1.

The relationship between the parameter $\alpha$ and puncture size was observed to be non-linear. To model this, we employed a non-linear regression approach using a single-layer perceptron with a tanh activation function. This model was trained on the extracted features from the IMU data samples and achieved a mean square error (MSE) of 0.0015. Fig.~\ref{fig:Severity Compare}(d) illustrates the first two principal components of the feature space, highlighting the clustered distribution of samples corresponding to different puncture severities.

\begin{figure}[h]
      \centering
      \includegraphics[width=1\linewidth]{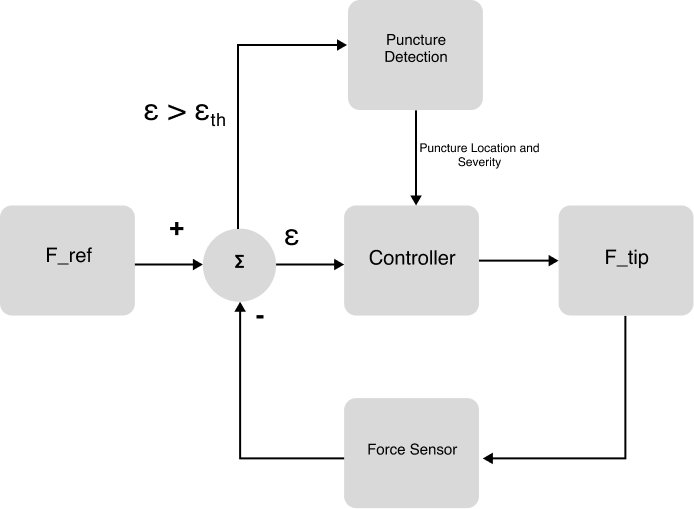}
      \caption{Failure recovery control schematic utilizing proposed puncture detection algorithm}
      \label{fig:failure_recovery}
   \end{figure}
   
\subsection{Puncture Failure Recovery}

In this section, we present the results of a puncture recovery strategy based on the proposed puncture detection methodology. Punctures in soft actuators can degrade performance and destabilize control; however, not all failures impact the actuator to the same extent. For instance, in the case of minor punctures (e.g., 0.51 mm), the actuator may still operate with reduced efficiency. In contrast, more severe damage can significantly impair functionality. In such cases, omitting the affected chamber from further actuation may improve overall system performance, efficiency, and controllability. 

We present a candidate control strategy designed to compensate for errors arising from high-severity punctures in multi-chamber soft actuators. In this framework, the actuator tip force is treated as the primary control variable and is directly impacted by punctures in the pneumatic chambers. The proposed control architecture is illustrated in Fig.~\ref{fig:failure_recovery}.

\begin{figure}[h]
      \centering
      \includegraphics[width=1\linewidth]{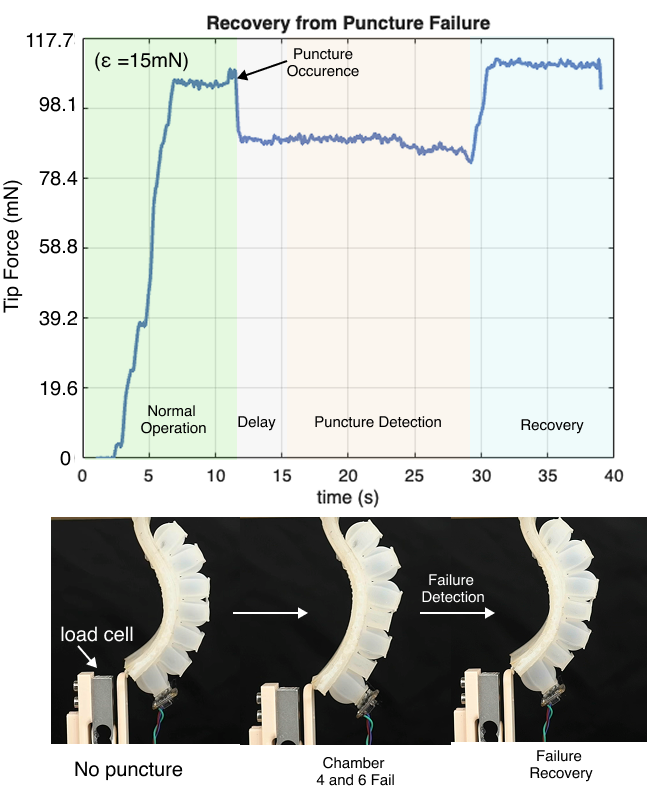}
      \caption{Failure recovery results}
      \label{fig:failure_recover_results}
   \end{figure}

During normal operation, all chambers are uniformly pressurized to $P_0$, enabling the actuator to generate the desired tip force $f_{\text{tip}}= f_{\text{ref}} $. However, when a failure occurs—such as in chambers 4 and 6—the actuator may no longer be able to achieve $f_{\text{ref}}$ under the same input $P_0$. (See Fig. \ref{fig:failure_recover_results}) A threshold parameter $\epsilon_{\text{th}}$ is used to trigger  puncture detection based on deviation from the desired tip force $f_{ref}$. To avoid transient disturbances immediately following a failure, a short delay is introduced before initiating the puncture detection process.

Once punctured chambers are identified, a decision is made—based on the severity of each puncture—whether to continue or exclude the affected chambers from further actuation. Empirical observations suggest that chambers with a severity score below 0.3 can continue operating without introducing significant instability to the system. Furthermore, the intact chambers are inflated equally until the desired tip force $f_{\text{ref}}$ is achieved. It should be noted that, depending on the number of failed chambers, such a recovery might not always be possible, and further investigation is required to evaluate its scope.

   
\section{Discussion}

This study presents a lightweight machine learning framework for detecting puncture failures and estimating their severity in pneumatic soft actuators. Our approach relies solely on inertial measurements from a single IMU, offering a sensor-efficient solution that is potentially generalizable across different pneumatic or even hydraulic soft robotic platforms. We introduce a chamber perturbation strategy that enables independent inference of both puncture location and severity across all actuator chambers. By targeting both localization and severity quantification, our method enables scalable, real-time failure detection to extend the lifetime of soft robots. These capabilities offer a foundation for failure-aware control strategies in soft robotic systems, enhancing robustness and reliability in practical deployments.

A key observation from our study is the variability in puncture localization accuracy across model architectures, particularly under unseen external loading conditions. While fully connected autoencoder models achieved up to 98\% accuracy in puncture localization—even under out-of-distribution scenarios—statistical methods such as OC-SVMs and isolation forests exhibited significant performance degradation when the actuator was subject to external perturbations not present in the training data. This highlights a critical trade-off: although SVMs and ensemble-based models offer simplicity and interpretability, they may lack robustness in generalizing to real-world variations. In contrast, autoencoder-based anomaly detection methods demonstrate stronger resilience, making them more suitable for real-time, multi-modal applications where robustness to unexpected disturbances is essential.

While these results are promising, further validation is necessary to assess the method’s scalability across diverse soft robotic designs and a wider range of external disturbances. This includes collecting data under varying conditions and retraining models where appropriate. Notably, feature extraction-based models maintained high accuracy under static external loads; however, their performance is expected to degrade under dynamic or time-varying loading conditions, which can obscure motion-dependent signatures associated with puncture events. For multi-segment soft robots, each segment can be perturbed to collect IMU data, with sensor placement chosen to reliably capture failure associated motion signatures across all segments.

Currently, our failure detection framework has been validated only on soft actuators fabricated using Ecoflex 00-50, a widely used silicone elastomer. As part of future work, we will evaluate the generalizability of our approach across actuators fabricated with different silicone elastomers—such as Ecoflex 00-35, 00-30, and even the more compliant 00-10—that are widely used in soft robotics. We also plan to explore the robustness of the current method under more complex scenarios involving multiple simultaneous punctures in different soft-actuator designs. We also aim to extend this evaluation to alternative silicone-based materials with diverse mechanical properties to further assess robustness across material variations. Furthermore, with growing interest in sustainable alternatives to conventional soft robots, we plan to extend our methodology to biodegradable yet mechanically fragile materials, such as gelatin-based actuators \cite{baumgartner2020resilient}. 
By enabling early failure detection in these systems, our approach can help extend functional lifespans while ensuring safe and predictable degradation, ultimately advancing the design of environmentally sustainable soft robotic platforms.

\section{CONCLUSIONS}

This paper presents a robust, in-situ method for puncture detection and severity estimation in pneumatic soft robots, leveraging only onboard IMU data collected via a chamber-perturbation strategy. Puncture localization is achieved using autoencoder-based models, while severity is estimated via a non-linear regression approach—together enabling failure-informed control decisions without the need for external instrumentation. Additionally, we propose a candidate recovery strategy for scenarios involving multi-chamber failures, illustrating the framework's practical utility. The results highlight the potential of IMU-based sensing for real-time, onboard failure detection, particularly in untethered soft robotic systems operating in unstructured or unpredictable environments.

The proposed method demonstrates strong scalability and cost-effectiveness, making it adaptable to a wide range of pneumatic soft actuator designs and materials. Its reliance on minimal sensing infrastructure enables potential extension to more sustainable alternatives in the future. 









{
\bibliographystyle{IEEEtran}
\bibliography{ref}

@article{baumgartner2020resilient,
  title={Resilient yet entirely degradable gelatin-based biogels for soft robots and electronics},
  author={Baumgartner, Melanie and Hartmann, Florian and Drack, Michael and Preninger, David and Wirthl, Daniela and Gerstmayr, Robert and Lehner, Lukas and Mao, Guoyong and Pruckner, Roland and Demchyshyn, Stepan and others},
  journal={Nature Materials},
  volume={19},
  number={10},
  pages={1102--1109},
  year={2020},
  publisher={Nature Publishing Group UK London}
}

@article{pan2014ultra,
  title={An ultra-sensitive resistive pressure sensor based on hollow-sphere microstructure induced elasticity in conducting polymer film},
  author={Pan, Lijia and Chortos, Alex and Yu, Guihua and Wang, Yaqun and Isaacson, Scott and Allen, Ranulfo and Shi, Yi and Dauskardt, Reinhold and Bao, Zhenan},
  journal={Nature communications},
  volume={5},
  number={1},
  pages={3002},
  year={2014},
  publisher={Nature Publishing Group UK London}
}

@article{mannsfeld2010highly,
  title={Highly sensitive flexible pressure sensors with microstructured rubber dielectric layers},
  author={Mannsfeld, Stefan CB and Tee, Benjamin CK and Stoltenberg, Randall M and Chen, Christopher V HH and Barman, Soumendra and Muir, Beinn VO and Sokolov, Anatoliy N and Reese, Colin and Bao, Zhenan},
  journal={Nature materials},
  volume={9},
  number={10},
  pages={859--864},
  year={2010},
  publisher={Nature Publishing Group UK London}
}

@article{yin2021wearable,
  title={Wearable soft technologies for haptic sensing and feedback},
  author={Yin, Jessica and Hinchet, Ronan and Shea, Herbert and Majidi, Carmel},
  journal={Advanced Functional Materials},
  volume={31},
  number={39},
  pages={2007428},
  year={2021},
  publisher={Wiley Online Library}
}

@article{li2021self,
  title={Self-powered soft robot in the Mariana Trench},
  author={Li, Guorui and Chen, Xiangping and Zhou, Fanghao and Liang, Yiming and Xiao, Youhua and Cao, Xunuo and Zhang, Zhen and Zhang, Mingqi and Wu, Baosheng and Yin, Shunyu and others},
  journal={Nature},
  volume={591},
  number={7848},
  pages={66--71},
  year={2021},
  publisher={Nature Publishing Group UK London}
}

@inproceedings{miyashita2016ingestible,
  title={Ingestible, controllable, and degradable origami robot for patching stomach wounds},
  author={Miyashita, Shuhei and Guitron, Steven and Yoshida, Kazuhiro and Li, Shuguang and Damian, Dana D and Rus, Daniela},
  booktitle={2016 IEEE international conference on robotics and automation (ICRA)},
  pages={909--916},
  year={2016},
  organization={IEEE}
}

@article{ramuz2012transparent,
  title={Transparent, optical, pressure-sensitive artificial skin for large-area stretchable electronics.},
  author={Ramuz, Marc and Tee, BC and Tok, JB and Bao, Zhenan},
  journal={Advanced materials (Deerfield Beach, Fla.)},
  volume={24},
  number={24},
  pages={3223--3227},
  year={2012}
}

@article{thuruthel_soft_2019,
	title = {Soft robot perception using embedded soft sensors and recurrent neural networks},
	volume = {4},
	url = {https://www.science.org/doi/full/10.1126/scirobotics.aav1488},
	doi = {10.1126/scirobotics.aav1488},
	number = {26},
	urldate = {2024-09-06},
	journal = {Science Robotics},
	author = {Thuruthel, Thomas George and Shih, Benjamin and Laschi, Cecilia and Tolley, Michael Thomas},
	month = jan,
	year = {2019},
	note = {Publisher: American Association for the Advancement of Science},
	pages = {eaav1488},
}

@Article{s140712748,
AUTHOR = {Culha, Utku and Nurzaman, Surya G. and Clemens, Frank and Iida, Fumiya},
TITLE = {SVAS3: Strain Vector Aided Sensorization of Soft Structures},
JOURNAL = {Sensors},
VOLUME = {14},
YEAR = {2014},
NUMBER = {7},
PAGES = {12748--12770},
URL = {https://www.mdpi.com/1424-8220/14/7/12748},
PubMedID = {25036332},
ISSN = {1424-8220},
DOI = {10.3390/s140712748}
}

@ARTICLE{8624340,
  author={Lin, Nan and Wu, Peichen and Wang, Menghao and Wei, Jizhou and Yang, Fan and Xu, Songlin and Ye, Zhicheng and Chen, Xiaoping},
  journal={IEEE Robotics and Automation Letters}, 
  title={IMU-Based Active Safe Control of a Variable Stiffness Soft Actuator}, 
  year={2019},
  volume={4},
  number={2},
  pages={1247-1254},
  doi={10.1109/LRA.2019.2894856}}

@INPROCEEDINGS{7139569,
  author={Santaera, Gaspare and Luberto, Emanuele and Serio, Alessandro and Gabiccini, Marco and Bicchi, Antonio},
  booktitle={2015 IEEE International Conference on Robotics and Automation (ICRA)}, 
  title={Low-cost, fast and accurate reconstruction of robotic and human postures via IMU measurements}, 
  year={2015},
  volume={},
  number={},
  pages={2728-2735},
  doi={10.1109/ICRA.2015.7139569}}

@Article{s22062205,
AUTHOR = {Webert, Heiko and Döß, Tamara and Kaupp, Lukas and Simons, Stephan},
TITLE = {Fault Handling in Industry 4.0: Definition, Process and Applications},
JOURNAL = {Sensors},
VOLUME = {22},
YEAR = {2022},
NUMBER = {6},
ARTICLE-NUMBER = {2205},
URL = {https://www.mdpi.com/1424-8220/22/6/2205},
PubMedID = {35336376},
ISSN = {1424-8220},
DOI = {10.3390/s22062205}
}

@article{SHAPIRO2011484,
title = {Bi-bellows: Pneumatic bending actuator},
journal = {Sensors and Actuators A: Physical},
volume = {167},
number = {2},
pages = {484-494},
year = {2011},
note = {Solid-State Sensors, Actuators and Microsystems Workshop},
issn = {0924-4247},
doi = {https://doi.org/10.1016/j.sna.2011.03.008},
url = {https://www.sciencedirect.com/science/article/pii/S092442471100135X},
author = {Yoel Shapiro and Alon Wolf and Kosa Gabor}
}

@ARTICLE{8693775,
  author={Wang, Jiangbei and Fei, Yanqiong and Pang, Wu},
  journal={IEEE/ASME Transactions on Mechatronics}, 
  title={Design, Modeling, and Testing of a Soft Pneumatic Glove With Segmented PneuNets Bending Actuators}, 
  year={2019},
  volume={24},
  number={3},
  pages={990-1001},
  doi={10.1109/TMECH.2019.2911992}}

@INPROCEEDINGS{11128570,
  author={Krings, Ethan J. and McManigal, Patrick and Markvicka, Eric J.},
  booktitle={2025 IEEE International Conference on Robotics and Automation (ICRA)}, 
  title={Intelligent Self-Healing Artificial Muscle: Mechanisms for Damage Detection and Autonomous Repair of Puncture Damage in Soft Robotics}, 
  year={2025},
  volume={},
  number={},
  pages={2591-2598},
  doi={10.1109/ICRA55743.2025.11128570}}

@article{Abidi2017,
  title = {On Intrinsic Safety of Soft Robots},
  volume = {4},
  ISSN = {2296-9144},
  url = {http://dx.doi.org/10.3389/frobt.2017.00005},
  DOI = {10.3389/frobt.2017.00005},
  journal = {Frontiers in Robotics and AI},
  publisher = {Frontiers Media SA},
  author = {Abidi,  Haider and Cianchetti,  Matteo},
  year = {2017},
  month = feb 
}

@ARTICLE{9072123,
  author={Sinaga, Kristina P. and Yang, Miin-Shen},
  journal={IEEE Access}, 
  title={Unsupervised K-Means Clustering Algorithm}, 
  year={2020},
  volume={8},
  number={},
  pages={80716-80727},
  doi={10.1109/ACCESS.2020.2988796}}

@INPROCEEDINGS{8101921,
  author={Flanagan, Kieran and Fallon, Enda and Connolly, Paul and Awad, Abir},
  booktitle={2017 Internet Technologies and Applications (ITA)}, 
  title={Network anomaly detection in time series using distance based outlier detection with cluster density analysis}, 
  year={2017},
  volume={},
  number={},
  pages={116-121},
  doi={10.1109/ITECHA.2017.8101921}}

@misc{Jimultimodal,
  doi = {10.48550/ARXIV.2012.08637},
  url = {https://arxiv.org/abs/2012.08637},
  author = {Ji,  Tianchen and Vuppala,  Sri Theja and Chowdhary,  Girish and Driggs-Campbell,  Katherine},
  title = {Multi-Modal Anomaly Detection for Unstructured and Uncertain Environments},
  publisher = {arXiv},
  year = {2020},
  copyright = {arXiv.org perpetual,  non-exclusive license}
}

@article{Barbado_2022,
   title={Rule extraction in unsupervised anomaly detection for model explainability: Application to OneClass SVM},
   volume={189},
   ISSN={0957-4174},
   url={http://dx.doi.org/10.1016/j.eswa.2021.116100},
   DOI={10.1016/j.eswa.2021.116100},
   journal={Expert Systems with Applications},
   publisher={Elsevier BV},
   author={Barbado, Alberto and Corcho, Óscar and Benjamins, Richard},
   year={2022},
   month=mar, pages={116100} }

@ARTICLE{8279425,
  author={Park, Daehyung and Hoshi, Yuuna and Kemp, Charles C.},
  journal={IEEE Robotics and Automation Letters}, 
  title={A Multimodal Anomaly Detector for Robot-Assisted Feeding Using an LSTM-Based Variational Autoencoder}, 
  year={2018},
  volume={3},
  number={3},
  pages={1544-1551},
  doi={10.1109/LRA.2018.2801475}}

@INPROCEEDINGS{9617498,
  author={Al Farizi, Wahid Salman and Hidayah, Indriana and Rizal, Muhammad Nur},
  booktitle={2021 8th International Conference on Information Technology, Computer and Electrical Engineering (ICITACEE)}, 
  title={Isolation Forest Based Anomaly Detection: A Systematic Literature Review}, 
  year={2021},
  volume={},
  number={},
  pages={118-122},
  doi={10.1109/ICITACEE53184.2021.9617498}}

@misc{alvarez2022revealinglargescaleevaluationunsupervised,
      title={A Revealing Large-Scale Evaluation of Unsupervised Anomaly Detection Algorithms}, 
      author={Maxime Alvarez and Jean-Charles Verdier and D'Jeff K. Nkashama and Marc Frappier and Pierre-Martin Tardif and Froduald Kabanza},
      year={2022},
      eprint={2204.09825},
      archivePrefix={arXiv},
      primaryClass={cs.LG},
      url={https://arxiv.org/abs/2204.09825}, 
}

@article{Qu2022,
  title = {Modeling nonlinear viscoelastic responses of flexible composites for soft robotics applications},
  volume = {30},
  ISSN = {1537-6532},
  url = {http://dx.doi.org/10.1080/15376494.2022.2063460},
  DOI = {10.1080/15376494.2022.2063460},
  number = {14},
  journal = {Mechanics of Advanced Materials and Structures},
  publisher = {Informa UK Limited},
  author = {Qu,  Jian and Khan,  Kamran and Meng,  Songhe and Muliana,  Anastasia},
  year = {2022},
  month = apr,
  pages = {2793–2805}
}

@Article{machines10060459,
AUTHOR = {Wang, Haiming and Li, Qiang and Liu, Yongqiang and Yang, Shaopu},
TITLE = {Anomaly Data Detection of Rolling Element Bearings Vibration Signal Based on Parameter Optimization Isolation Forest},
JOURNAL = {Machines},
VOLUME = {10},
YEAR = {2022},
NUMBER = {6},
ARTICLE-NUMBER = {459},
URL = {https://www.mdpi.com/2075-1702/10/6/459},
ISSN = {2075-1702},
DOI = {10.3390/machines10060459}
}

@article{Fidali2024,
  title = {Evaluation of the Diagnostic Sensitivity of Digital Vibration Sensors Based on Capacitive MEMS Accelerometers},
  volume = {24},
  ISSN = {1424-8220},
  url = {http://dx.doi.org/10.3390/s24144463},
  DOI = {10.3390/s24144463},
  number = {14},
  journal = {Sensors},
  publisher = {MDPI AG},
  author = {Fidali,  Marek and Augustyn,  Damian and Ochmann,  Jakub and Uchman,  Wojciech},
  year = {2024},
  month = jul,
  pages = {4463}
}

@INPROCEEDINGS{9144922,
  author={Shao, Qiwen and Zhang, Ningbin and Shen, Zequn and Gu, Guoying},
  booktitle={2020 17th International Conference on Ubiquitous Robots (UR)}, 
  title={A Pneumatic Soft Gripper with Configurable Workspace and Self-sensing}, 
  year={2020},
  volume={},
  number={},
  pages={36-43},
  doi={10.1109/UR49135.2020.9144922}}

@article{PneumaticMuscle,
author = {Daerden, Frank and Lefeber, Dirk},
year = {2002},
month = {03},
pages = {},
title = {Pneumatic Artificial Muscles: actuators for robotics and automation},
volume = {47},
journal = {European Journal of Mechanical and Environmental Engineering}
}
}

\end{document}